\documentclass[conference]{IEEEtran}
\usepackage{times}

\usepackage[numbers]{natbib}
\usepackage{multicol}
\usepackage{xcolor}
\usepackage[bookmarks=true,hidelinks]{hyperref}

\usepackage[linesnumbered,ruled,vlined]{algorithm2e}
\usepackage{graphicx}
\usepackage{mathtools}
\usepackage{amssymb}
\usepackage{subcaption}
\newcommand\blfootnote[1]{%
  \begingroup
  \renewcommand\thefootnote{}\footnote{#1}%
  \addtocounter{footnote}{-1}%
  \endgroup
}

\DeclareCaptionLabelSeparator{periodspace}{.\quad}
\begin{document}
\title{Concurrent Constrained Optimization of Unknown Rewards for Multi-Robot Task Allocation}


\author{\authorblockN{Sukriti Singh{$^\dagger$}, Anusha Srikanthan{$^{\ddagger}$}, Vivek Mallampati{$^\dagger$}, Harish Ravichandar{$^\dagger$}} 
\authorblockA{{$^\dagger$}Georgia Institute of Technology, {$^\ddagger$}University of Pennsylvania\\
Email: \{sukriti, vmallampati6, harish.ravichandar\}@gatech.edu, sanusha@seas.upenn.edu}
}


%

\maketitle

\begin{abstract}
Task allocation can enable effective coordination of multi-robot teams to accomplish tasks that are intractable for individual robots. However, existing approaches to task allocation often assume that task requirements or reward functions are known and explicitly specified by the user. In this work, we consider the challenge of forming effective coalitions for a given heterogeneous multi-robot team when task reward functions are \textit{unknown}. To this end, we first formulate a new class of problems, dubbed \textit{COncurrent Constrained Online optimization of Allocation (COCOA)}. The COCOA problem requires online optimization of coalitions such that the unknown rewards of all the tasks are \textit{simultaneously} maximized using a given multi-robot team with \textit{constrained} resources. To address the COCOA problem, we introduce an online optimization algorithm, named \textit{Concurrent Multi-Task Adaptive Bandits (CMTAB)}, that leverages and builds upon continuum-armed bandit algorithms. Experiments involving detailed numerical simulations and a simulated emergency response task reveal that CMTAB can effectively trade-off exploration and exploitation to simultaneously and efficiently optimize the unknown task rewards while respecting the team's resource constraints.\footnote{Source code and appendices are available at \url{https://github.com/GT-STAR-Lab/CMTAB}.}\blfootnote{*This work was supported by the Army Research Lab under Grant W911NF-20-2-0036.}
\end{abstract}

\IEEEpeerreviewmaketitle

\section{Introduction}
Complex real-world challenges require the coordination of heterogeneous robots. Instances like these arise in several domains, such as search and rescue operations \cite{Zhao22,Grabowski00}, industrial processes \cite{Naidoo16}, and multi-scale environmental monitoring \cite{Salam21,Micael19}. Indeed, the effectiveness of coordination among robots within such heterogeneous teams has a direct impact on their success. One form of such coordination involves the challenge of careful allocation of tasks to agents in order to form suitable coalitions. In fact, this is a well-studied challenge that is referred to as multi-robot task allocation (MRTA)~\cite{gerkey2004formal}. 

Existing approaches that form coalitions in heterogeneous teams often assume that task requirements, reward functions, or utilities are provided apriori by the user. However, complex real-world problems seldom allow for such explicit specifications. Indeed, it is known that humans struggle to explicitly specify complex requirements~\cite{rieskamp2003people}.

In this work, we undertake the challenge of forming effective coalitions for a given multi-robot team when task reward or utility functions are \textit{unknown}.
Specifically, we formulate and solve a new class of problems that involve optimizing coalitions for a given team such that the \textit{unknown} reward functions associated with the tasks are \textit{simultaneously} maximized in as few iterations as possible. We call such problems \textit{COncurrent Constrained Online optimization of Allocation (COCOA)}. The need to maximize rewards in as few interactions as possible is motivated by the fact that ``trying out" allocations can be prohibitively expensive in multi-robot settings either due to deployment costs or the computational burden of running a large number of simulations. 

Given that we are interested in the online optimization of unknown functions, the COCOA problem has strong connections to the rich literature in Bayesian online optimization. However, unlike typical Bayesian optimization problems, COCOA involves two challenges that are unique to multi-robot task allocation: i) \textit{concurrency}: COCOA requires that all tasks be optimized concurrently since the tasks are likely to belong to a unified mission, and ii) \textit{resource constraints}: any given team will have a limited number of agents and associated traits (i.e., capabilities such as speed, sensing radius, and payload). As such, one cannot allocate arbitrary amounts of robots or traits. Further, robot traits are indivisible (e.g., one cannot utilize a robot's payload in one task and its speed in another.). 

To solve the COCOA problem, we contribute a novel algorithm that we call \textit{Concurrent Multi-Task Adaptive Bandits (CMTAB)}. A straightforward approach to learning to optimize coalitions when task reward functions are unknown would be to learn the ideal allocation of agents to tasks such that task rewards are maximized. However, a downside to such an approach is that the learned model will not generalize to teams comprised of new agents and will require retraining. In contrast, CMTAB models the reward function for each task as a \textit{trait-reward map}. These maps enable generalization by capturing how task reward varies as a function of the collective multi-dimensional traits allocated to that task (e.g., total payload capacity, aggregated sensing range, etc.).

\begin{figure}
    \centering    \includegraphics[width=0.9\columnwidth]{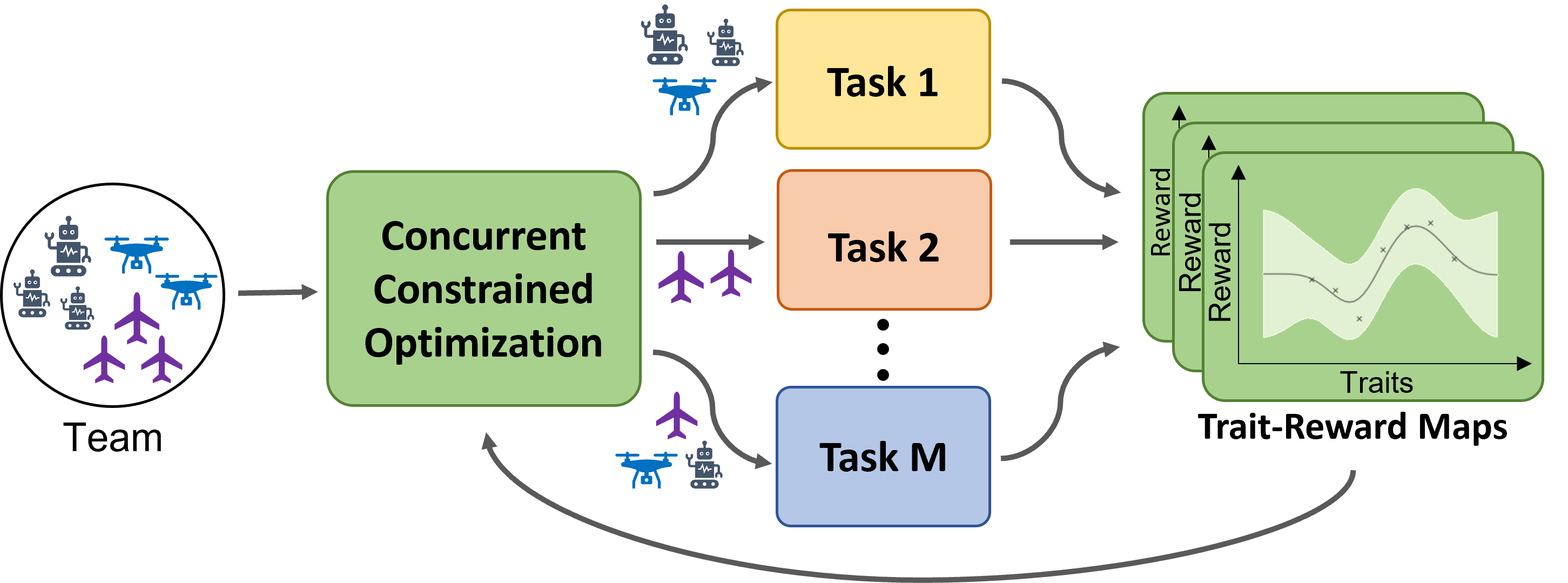}
    \caption{\small{An illustration of our approach to \textit{online} optimization of task allocation under resource constraints in order to \textit{concurrently} maximize the \textit{unknown} reward functions of multiple tasks.}}
    \label{fig:block_diagram}
\end{figure}

CMTAB leverages ideas from multi-arm-bandit-based approaches in order to maximize unknown task rewards efficiently. Bandit-based algorithms are a natural fit for the COCOA problem as they strive to maximize unknown reward functions (i.e., exploitation) while simultaneously minimizing the number of suboptimal decisions (i.e., the cost of exploration). Given that CMTAB models reward functions as trait-reward maps, the action or input space for each reward function is continuous due to the fact that robot traits belong to a continuous space. CMTAB leverages the seminal work on Gaussian Process (GP) bandits~\cite{srinivas2009gaussian} to efficiently optimize over the continuous trait space.

CMTAB builds upon the original GP-Upper Confidence Bound (GP-UCB) algorithm \cite{srinivas2009gaussian} in a number of ways to handle challenges unique to the COCOA problem. First, CMTAB \textit{simultaneously} runs multiple instances of GP-UCB (one for each task) in order to handle COCOA's concurrency challenge. Next, when selecting coalitions to sample for each task, CMTAB must respect the restrictions imposed by the \textit{finite} resources of the team (i.e., robots and traits). Finally, CMTAB adopts an adaptive discretization technique to effectively sample from the high-dimensional continuous trait space. Inspired by the zooming algorithm~\cite{Slivkins19}, CMTAB adaptively prioritizes finer discretization of high-reward regions.

In addition to learning via reinforcement, CMTAB can leverage historical data (i.e., passive demonstrations of allocations and associated rewards). 
CMTAB does not attempt to directly imitate offline data and instead uses them to initialize the trait-reward maps before engaging in online optimization. As such, CMTAB does not require the demonstrations to be optimal with respect to task rewards. To the best of our knowledge, there has been no prior work on learning coalition formation from demonstrations of varying quality.

In summary, we contribute 
\begin{itemize}
    \item COCOA: A novel problem formulation for online optimization of heterogeneous coalition formation under resource constraints to concurrently maximize \textit{unknown} task reward functions.
    \item CMTAB: A Bayesian online optimization algorithm capable of solving the COCOA problem.
    \item An approach to incorporate offline historical data exhibiting varying performance to bootstrap online optimization of coalition formation. 
\end{itemize}

We evaluate CMTAB using detailed numerical simulations and a simulated emergency response mission, and compare it against ablative baselines to validate our design choices. Our results demonstrate that CMTAB consistently and considerably outperforms all baselines in terms of task performance, sample efficiency, and cumulative regret. 

\section{Related Work}

In this section, we situate our contributions in relevant sub-fields to highlight the similarities and differences between our work and prior efforts from different vantage points. 

\vspace{3pt}
\noindent \textbf{Multi-robot Task Allocation}:  We focus on the Single-Task  robots, Multi-Robot tasks, and Instantaneous Allocation (ST-MR-IA) version of MRTA~\cite{gerkey2004formal,korsah2013comprehensive}. Solutions to the ST-MR-IA problem can be categorized into one of the following groups: market-based methods~\cite{guerrero2003multi,lin2005combinatorial,xie2018mutual}, numerical-optimization-based methods~\cite{notomista2019optimal}, and the recently introduced trait-based methods~\cite{prorok2017impact,ravichandar2020recent,messing2022grstaps}. Our work falls into the category of trait-based methods, 
but differs from existing approaches in two important ways. First, all existing trait-based approaches assume a binary model (success/failure) to task performance. In contrast, our approach adopts a richer model of task performance in the form of reward functions that continuously vary with allocated traits, recognizing the fact that real-world scenarios admit a spectrum of performance.
Second, most existing trait-based approaches assume that task requirements are explicitly specified by the user in terms of traits. In contrast, we do not assume that these reward functions are known and develop an efficient way to optimize these unknown reward functions online under resource constraints.

\vspace{3pt}
\noindent \textbf{Multi-Arm Bandits}:
Our problem formulation and solution are intricately connected to the rich literature of multi-arm bandit problems. Unlike traditional multi-arm bandit problems, our problem formulation belongs to the category of continuum-arm bandits in which the action space is continuous. Our approach CMTAB is inspired by two popular techniques for solving continuum-arm bandit problems. First, CMTAB builds upon GP-UCB~\cite{srinivas2009gaussian} algorithm to model the trait-reward maps for each task using a GP. Second, CMTAB's adaptive discretization algorithm is inspired from the Zooming algorithm~\cite{Slivkins19} to adaptively discretize continuous action spaces such that, high reward regions are sampled more often than other regions. Indeed, the use of these approaches in multi-robot task allocation is by itself novel. However, one cannot merely apply these two approaches without modification to solve the proposed COCOA problem. 
This is due to 
a combination of challenges posed by multi-robot task allocation problems, typically not found in the multi-arm bandit literature (i.e. task concurrence, and finite and indivisible robot traits)
In essence, CMTAB involves a novel extension of these ideas in order to concurrently maximize the unknown reward functions of multi tasks by optimizing the allocation of a heterogeneous multi-robot team.

\vspace{3pt}
\noindent \textbf{Interaction-based learning for Allocation}:
The proposed approach is most closely related to approaches that employ interaction-based learning (e.g., multi-arm bandits) to solve resource allocation problems. 
Multi-agent task assignment in the bandit framework~\cite{le2006multi} proposed the formulation of task allocation as a restless bandit problem with switching costs and discounted rewards. 
More recent work~\cite{rangi2018multi} addresses issues in crowdsourcing systems by estimating a worker's ability over time in a multi-armed bandit setup to improve the performance on allocated tasks. Prior work has also leveraged multi-armed bandits for multi-agent network optimization by using local interactions to improving the network performance~\cite{shahrampour2017multi, wang2010evolutionary}.
However, a common assumption made by these approaches is the access to virtually unlimited resources while learning to allocate resources. 
In stark contrast, our approach explicitly accounts for the fact that multi-robot task allocation imposes unavoidable resource constraints due to finite and indivisible robot traits. 

\vspace{3pt}
\noindent \textbf{Multi-task Learning}:
Prior efforts have also investigated optimization of multiple unknown objectives and multi-task learning. \citet{Bonilla07} developed a framework to model and predict the outputs of multiple tasks. However, they assume that each task is marginally identically distributed up to a scaling factor.
\citet{Sener18} apply gradient-based multi-objective optimization to multi-task learning when tasks have conflicting objectives with shared parameters.
There is also a rich body of work in multi-objective multi-armed bandit problems~\cite{Roijers17,Wan21,Louedec15,Dai20,Pearce18} that typically exploit inter-objective or inter-task dependencies to optimize multiple objectives.
However, these multi-task learning methods address problems that lack a key challenge present in our approach: concurrence. Typically, these approaches enjoy the luxury of individually and independently sampling arms for each task. In contrast, our approach learns to optimize the allocation of heterogeneous robots to tasks that require concurrent execution without assuming access to any side information.

\vspace{3pt}
\noindent \textbf{Summary}: The various challenges related to our work (e.g., online optimization, continuous and constrained inputs, multi-task learning) have indeed been individually explored extensively in the literature. However, our setting of online optimization of heterogeneous multi-robot coalition formation under resource constraints presents a unique combination of these challenges. Hence, we contribute both a new problem formulation named \textit{COncurrent Constrained Online optimization of Allocation (COCOA)} and a first solution named Concurrent Multi-Task Adaptive Bandit (CMTAB). \textcolor{black}{Note that given its novelty, COCOA cannot be solved using existing methods without significant modification. Therefore, our CMTAB algorithm is not a contender among state-of-the-art algorithms in MRTA, but rather the first attempt at tackling COCOA.}

\section{Problem Formulation}
\label{sec:problem}
In this section, we formulate a new class of active learning problems for heterogeneous task allocation, that we term COncurrent Constrained Online optimization of Allocation (COCOA).
We consider a heterogeneous team in which each robot belongs to one of $S \in$ $\mathbb{N}$ \textit{species}, and $N_s$ number of robots in each species where $s \in \{1, 2,..., S\}$. 
Each species is categorized by $U \in$ $\mathbb{N}$ \textit{traits} (or capabilities) that are possessed by member robots.  
The capabilities of all robots in the team can be defined in a \textit{species-trait} matrix: $Q \in$ $\mathbb{R}^{SXU}_+$ whose $su$th element denotes the $u$th trait of a robot in the $s$th species.

We focus on the ST-MR-IA variant of task allocation~\cite{gerkey2004formal,korsah2013comprehensive}, in which the robots need to be allocated to $M$ independent but concurrent tasks,
such that each robot is only assigned to a single task. We denote the allocation using the \textit{assignment} matrix: $X \in$ $\mathbb{Z}^{MXS}_+$, where $x_{ms}$ denotes the number of agents from species $s$ assigned to the $m$th task. 

\textit{Trait aggregation}: For a given assignment $X$ of a team with the species-trait matrix $Q$, we compute the collective traits allocated to all the tasks as a \textit{task-trait} matrix $Y \in$  $\mathbb{R}^{MXU}_+$~\cite{ravichandar2020strata}:
\begin{equation} \label{eq:tasktrait}
    Y = X Q
\end{equation}
where each row is given by $y_m = Q^T x_m \in \mathcal{Y},\ \forall m$, and $\mathcal{Y} \subseteq \mathbb{R}^U$ is the space of all possible trait aggregation vectors.

\textit{Trait-reward maps}: We use $r_m$ to denote the noisy reward or reward obtained from the environment and model it as a function of $y_m$: 
\begin{equation}\label{eq:groundtruthreward}
    r_m(y_m) = f_m(y_m) + \epsilon_m
\end{equation}
where $\ f_m \colon\  \mathcal{Y}  \longrightarrow R_+$ is an unknown ground-truth \textit{trait-reward map} associated with the $m$th task that encapsulates the task reward as a function of the traits allocated to it, and $\epsilon_m$ captures the effects of 
inherent stochasticity in multi-robot tasks (e.g., sensing, actuation, environment, etc.) on task reward.

Note that we model the reward $r_m(\cdot)$ as a function of the aggregated capabilities $y_m$ and not the allocation of robots $x_m$. This choice is deliberate and motivated by the fact that models that map allocations to rewards are restricted to a specific team and are brittle to changes in team composition. 
In contrast, our model maps traits to rewards and can generalize to new teams, making it possible to learn \textit{team-agnostic} reward functions.

\vspace{3pt}
\noindent \textbf{Problem statement}: The overall objective of COCOA is to allocate robots from a given team such that the sum of unknown task rewards $r_{total}([y_1,\cdots,y_m])=\sum_m r_m(y_m)$ is maximized as quickly as possible. Formally, given a team with the species-trait matrix $Q$ and a set of $M$ tasks, we aim to learn to maximize the unknown total reward $r_{total}$ in as few interactions with the environment as possible. Indeed, there are two key challenges associated with COCOA:
\begin{itemize}
    \item \textit{Resource constraints}: We can only sample trait aggregations that are achievable by the given team.
    \item \textit{Concurrency}: We cannot individually sample tasks since the tasks have to be carried out concurrently.
\end{itemize}

We are particularly interested in the convergence of our optimization 
in as few interactions as possible, given that function evaluations (i.e.,``trying out" allocations) tend to be expensive either due to deployment costs associated with multi-robot systems or the complexity of running a large number of simulations. Further, COCOA prioritizes maximizing unknown rewards (i.e., exploration in the interest of exploitation) over accurately learning the trait-reward maps (i.e., pure exploration). See \citet{srinivas2009gaussian} for a related discussion of the connections between Bayesian optimization and experimental design.

\textit{Bootstrapped COCOA}: In addition to the above-defined COCOA, we also consider a second variant in which offline data $\mathcal{D} = \{X^{(n)},\ Q^{(n)},\ r_{total}^{(n)}\}_{n=1}^{N_{\mathcal{D}}}$ (consisting of allocations $X^{(n)}$ of teams with traits $Q^{(n)}$, and corresponding rewards $r^{(n)}$) are available. We refer to such data as \textit{demonstrations}. Indeed, in many multi-robot settings, it is likely that historical data from prior (manual) mission executions are available. A key challenge in solving this bootstrapped version of COCOA is that demonstrations may be both suboptimal (w.r.t. $r_{total}$) and come from teams that are different in composition than the one that is available for online learning as described above. 

\section{Online Optimization of Allocation under Resource Constraints}
In this section, we introduce our approach, named Concurrent Multi-Task Adaptive Bandit (CMTAB), to solve the COCOA problem introduced in Section \ref{sec:problem}. The terms concurrent and multi-task stem from the need to optimize the reward functions for all tasks simultaneously. CMTAB is adaptive because it leverages adaptive active sampling when optimizing unknown functions in high-dimensional continuous trait spaces. 

\subsection{Modeling Trait-Reward Maps as GPs}
We begin by modeling each of the $M$ ground-truth reward functions $\{f_m(\cdot)\}_{m=1}^M$ as a sample from a Gaussian Process (GPs)~\cite{srinivas2009gaussian} and then develop a bandit-based active learning algorithm to learn all GPs concurrently while respecting resource constraints of the team.
We choose GPs given their ability to model \textcolor{black}{a wide range of} continuous stochastic functions and effectiveness in the absence of metadata about the environment or task-specific features~\cite{Wan21}. Further, their non-parametric property does not require a hand-crafted heuristic. \textcolor{black}{Consequently, any reward function that can be represented using a GP is compatible with our methodology.}

Formally, we model the reward function of the $m$th task as sample from a GP: $f_m$ $\sim$ $\mathcal{GP}(\mu_m(y_m),k_m(y_m,\bar{y}_m))$
with its mean and covariance defined as
\begin{equation}
\mu_m(y_m) = \mathbb{E}[f_m(y_m)],\ \forall\  y_m \in \mathcal{Y}\\
\end{equation}
\begin{equation}
    k_m(y_m, \Bar{y}_m) = \mathbb{E}[(f_m(y_m)-\mu_m(y_m))  (f_m(\bar{y}_m)-\mu_m(\bar{y}_m))] \nonumber   
\end{equation}
where $y_m, \overline{y}_m \in \mathcal{Y}$ are two random variables in the trait vector space. 
To avoid numerical issues resulting from magnitude differences across traits, we normalize each trait with respect to the maximum magnitude of that trait that can possibly be assigned to any task from the given team.

We initialize each GP with a \textcolor{black}{Radial Basis Function} (RBF) kernel. At any iteration $ i \in \{1, \cdots, N\}$, the reward for the $m$th task ($r_m^{i}$) will be a function of the aggregated traits allocated to it ($y_m^{i}$) and is given by
$r_m^{i} = f_m(y_m^{i}) + \epsilon_m^{i}$,
where $\epsilon_m^{i} \sim \mathcal{N}(0,\sigma^2)$ denotes the i.i.d. noise, and $N$ is the number of iterations. 
After $N$ iterations of allocating traits $\textbf{Y}_m^{N} = \{y_m^{1},\ y_m^{2}, \cdots,\ y_m^{N}\}$ and collecting corresponding rewards' samples $\textbf{r}_m^{N} = [r_m^{1},\ r_m^{2}, \cdots,\ r_m^{N}]^T$, the posterior distribution of the reward function will still be a GP and its mean $\mu_m^{N}(y_m)$, covariance $k_m^{N}(y_m, \overline{y_m})$ and variance ${\sigma^2}_m^{N}(y_m)$ can be computed analytically as follows:

\begin{align}
    \mu_m^{N}(y_m) &= \textbf{k}_m^{N}(y_m)^T [\textbf{K}_m^{N}]^{-1}\textbf{r}_m^{N}, \nonumber \\
    k_m^{N}(y_m, \overline{y_m}) &= k_m(y_m,\ \overline{y}_m) - \textbf{k}_m^{N}(y_m)^T [\textbf{K}_m^{N}]^{-1} \textbf{k}_m^{N}(\overline{y_m}), \nonumber \\
    {\sigma^2}_m^{N}(y_m) &= k_m^{N}(y_m, y_m), \nonumber
\end{align}
where $\textbf{k}_m^{N}(y_m) = [ k_m(y_m^{1}, y_m),\ k_m(y_m^{2}, y_m), \cdots,\ \\k_m(y_m^{N}, y_m)]^T$ 
and $\textbf{K}_m^{N}$ is the positive definite kernel matrix $[k_m(y_m, \overline{y_m})]_{y_m,\ \overline{y_m} \in \textbf{Y}_m^{N}}$.

\subsection{Bandit-based Concurrent Constrained Online Optimization}\label{subsec:CMTAB}

With unknown trait-reward maps modeled as GPs, we now turn to the challenge of online optimization of the total task reward ($r_{total}([y_1,\cdots,y_m])=\sum_m r_m(y_m)$). To this end, CMTAB models the optimization of each task reward as a multi-armed bandit problem since bandit-based approaches offer a natural mechanism to optimize unknown reward functions in a sample-efficient manner~\cite{Slivkins19}.  The term ``arms" refer to the inputs of the unknown objective being optimized. In our work, the arms of the bandit associated with the $m$th task would represent the trait aggregations $y_m$, which serve as the input to the unknown trait-reward map. 
Note that each bandit has an infinite number of arms since trait aggregations $y_m$ belong to a continuous space ($\mathcal{Y}$), making each bandit a \textit{continuum-armed bandit}~\cite{Agrawal95}. The primary objective of CMTAB is to collectively sample arms for each of the $M$ bandits, such that the total reward ($r_{total}$) is maximized in as few iterations as possible. 

Since we have continuum-armed bandits, the trait space $\mathcal{Y}$ needs to be discretized. 
Unfortunately, fixed discretization 
methods are likely to fall prey to two failure modes in our setting: i) limited exploration power due to large discretization intervals, and ii) significant computational burden arising from small discretization intervals. 
To circumvent above-mentioned issues, CMTAB undertakes an \textit{adaptive discretization} approach called the zooming algorithm~\cite{Slivkins19} to identify and prioritize high-reward regions for finer discretization and hence exploitation. In addition to avoiding the pitfalls of fixed discretization, adaptive discretization also benefits from guaranteed bounds on regret~\cite{Slivkins19}. 

CMTAB differs from and builds upon standard techniques used in multi-armed bandit problems. Specifically, CMTAB solves multiple bandit problems \textit{concurrently} by sampling over multiple tasks. Further, CMTAB also ensures that the sampled point (i.e., a specific trait aggregation) is achievable by the given team's \textit{resource constraints} in terms of traits. Indeed, the challenges of adaptive discretization, concurrent sampling, and resource constraints are intertwined. Below, we describe how CMTAB optimizes the total reward by efficiently sampling allocations such that all three challenges are addressed (see Algorithm \ref{alg:CMTAB} for a pseudocode).

\begin{algorithm}
  \SetNoFillComment   
  \SetKwInOut{Input}{Input}
  \SetKwInOut{Output}{Output}
  \SetKwProg{try}{try}{:}{}
  \SetKwProg{catch}{except}{:}{end}
  \Input{$r_m^{i-1}\ \forall\ m \in [1,M],\ \mathbf{Y}_D$}
  \Output{$X^{i}$} 
  
    \For{\texttt{each $Y$ in $\mathbf{Y}_D$}}{
        Compute confidence radius, $\gamma^i(Y)$ as shown in (\ref{eq:confrad})\\
        Choose $N_f$ points in $Y$'s neighbourhood\\
        Compute estimated utility, $\zeta^i(Y)$ as shown in (\ref{eq:exp_reward})\\}
    Select $Y^i$ as shown in (\ref{eq:select-Yi})\\
    \try  {\texttt{Trait-satisfying allocation}}{
            Compute $X^i$ as shown in (\ref{eq:opt})\\
            with constraints: (\ref{eq:const1}),(\ref{eq:const2})}
    \catch{\texttt{Closest allocation}} {
            Compute $X^i$ as shown in (\ref{eq:opt})\\
            with constraint: (\ref{eq:const1})
           }

  \Return $X^i$
  \caption{Concurrent coalition selection at iteration $i$}
  \label{alg:CMTAB}
\end{algorithm}

We begin by coarsely discretizing the trait space $\mathcal{Y}$ into a collection of points denoted by $\mathcal{Y}_{D}$. 
Note that any collection of $M$ points in $\mathcal{Y}_{D}$ represents an instance of the aggregated traits for all $M$ tasks (denoted by the task-trait matrix $Y$ in Eq. \ref{eq:tasktrait}). 
We use $\mathbf{Y}_D$ to denote the set of all possible such combination of $M$ points in $\mathcal{Y}_{D}$. 
\textcolor{black}{If there are $d$ discretized intervals, then $\vert Y_D \vert \leq (d^U)^M$ since certain combinations would be eliminated due to the fact a team with finite resources cannot allocate certain distribution of capabilities.}
At every iteration, CMTAB selects an element of $\mathbf{Y}_D$ to achieve concurrent sampling based on two factors: i) confidence radius, and ii) estimated utility.

CMTAB assigns each element of $\mathbf{Y}_D$ a confidence radius at the $i$th iteration denoted by $\gamma^i(Y)$ and computed as follows
\begin{equation}\label{eq:confrad}
    \gamma^i(Y) = \sqrt{2\ log\ N/ (S_Y^{i-1} +1)}, \forall\ Y \in \mathbf{Y}_D
\end{equation}
where $N$ denotes the total number of iterations and $S_Y^{i-1}$ is the number of times the task-trait matrix $Y$ has been sampled in $i-1$ iterations.

CMTAB determines the estimated utility $\zeta^i(Y)$ of each $Y \in \mathbf{Y}_D$ at iteration $i$ by computing the average over $N_f$ points ($\{^j Y\}_{j=1}^{N_f}$) in its neighborhood as defined by its confidence radius $\gamma_i(Y)$:
\begin{equation}\label{eq:exp_reward}
    \zeta^i(Y) = \frac{1}{N_f}\sum_{j=1}^{N_f} \sum_{m=1}^{M} \mu^{(i-1)}_m(^j y_{m}) + \beta^{i} \sigma^{(i-1)}_m(^j y_{m})
\end{equation}
where $^j y_{m}$ denotes the aggregated traits allocated to the $m$th task if $^j Y$ is the task-trait matrix, and $\beta^{i}$ is a hyperparameter that trades off exploration (maximizing variance) and exploitation (maximizing expectation) as done in upper confidence bounds (UCB)~\cite{srinivas2009gaussian}. In order to respect the resource constraints imposed by the given team, CMTAB ensures that all $N_f$ points chosen in the neighborhood of $Y$ represent task-trait matrices that can be achieved by the team. Since all traits are normalized, this can be achieved by ensuring that each $Y$ satisfies $\sum_m y_{mu} \leq 1, \forall u \in \{1,\cdots,U\}$ where $y_{mu}$ is the $mu$th element of $Y$ representing the amount of $u$th trait allocated to the $m$th task. 

Note that smaller confidence intervals result in denser points closer to the task-trait matrix under consideration. Indeed, this ``rich get richer" mechanism is at the root of zooming algorithms, resulting in denser sampling in high-reward regions.

With both the estimated value $\zeta^i(Y)$ and confidence radius $\gamma^i(Y)$ computed, we can now define the UCB-based \textit{selection rule}~\cite{Slivkins19,srinivas2009gaussian} as follows
\begin{equation}\label{eq:select-Yi}
    Y^i = \arg \max_{Y \in \mathbf{Y}_D} \zeta^i(Y) + M \gamma^i(Y)
\end{equation}
We can further select a specific point among the $N_f$ feasible task-trait matrices in the neighborhood of $Y^i$ by selecting the one with the best 
estimated utility as defined in Eq. (\ref{eq:exp_reward}).

The above selection rule addresses the fundamental trade-off between exploration and exploitation. If the $\mu$'s (expected values) of the points in the neighborhood of a given combination of trait aggregation (i.e., task-trait matrix) is large,  we \textit{exploit} it. If that combination has not been selected yet or selected very few times, the $\sigma$'s (standard deviations) and confidence radius will make its utility larger so that it is \textit{explored}.

\textit{Selecting an allocation}: With this selection of a task-trait matrix as described above, we are left with identifying a specific allocation of robots $X$ that we can deploy. To this end, CMTAB solves the following constrained optimization problem to get an assignment of agents.

\begin{align}
    X^i = \arg \min_X\ & \Vert XQ - Y^i \Vert_2 \label{eq:opt}\\
    \mathrm{s.t.}\quad & \sum_{m=1} x_{ms} \leq N_s\ \forall\ s \in [1,S] \label{eq:const1} \\
                       & XQ \succeq Y^i \label{eq:const2}
\end{align}

\textcolor{black}{Initially, CMTAB attempts to obtain an assignment that strictly satisfies the target trait requirement $Y^i$ (Eq. (\ref{eq:const2})). If such an assignment is infeasible for a given team, CMTAB drops the constraint in Eq. (\ref{eq:const2}) and computes an assignment that minimizes the gap between the target trait aggregation $Y^i$ and the actual trait aggregation $XQ$.} This coalition $X^i$ is used to carry out the tasks in the environment at iteration $i$. The Gaussian process regression models of the reward functions are updated from rewards obtained by the coalition.

\textcolor{black}{Since we learn trait-reward maps, the problem's dimensionality doesn't increase with the number of robots ($N$) or species ($S$), and therefore doesn't add computational burden when selecting arms (lines 1-5 in Algorithm \ref{alg:CMTAB}). However, when $N$ and $S$ increase, coalition optimization (lines 6-11 of Algorithm \ref{alg:CMTAB}) will naturally incur a higher computational cost.}

\begin{figure*}
    \centering
    \includegraphics[width=\textwidth]{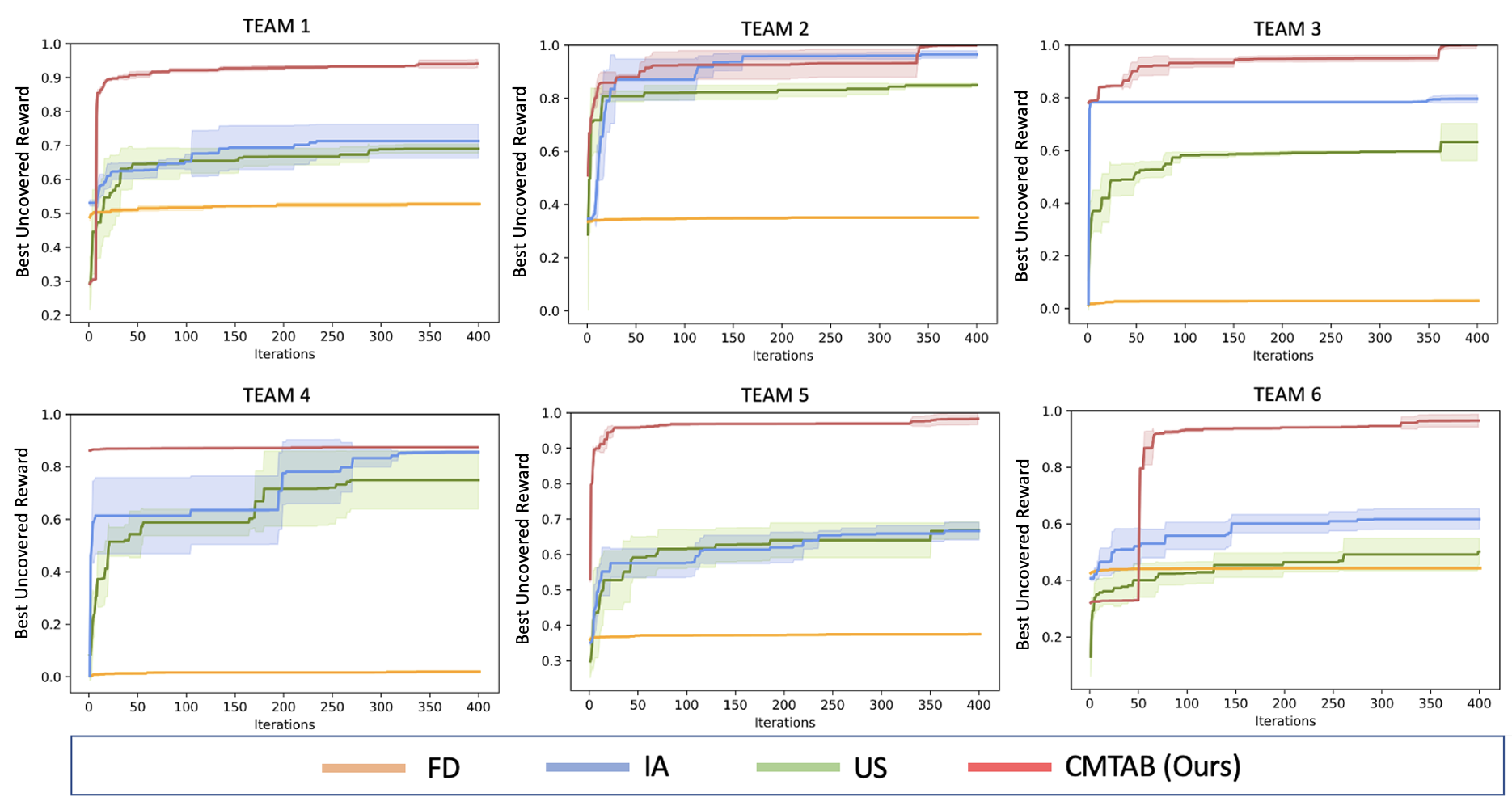}
    \caption{\small{Best uncovered reward as a function of interactions with the environment, when the GPs are initialized with zero mean.}}
    \label{fig:no_prior}
\end{figure*}

\subsection{Bootstrapping with Offline Data}
\label{subsec:bootstrapping}

In many realistic settings, the set of tasks and reward functions remain the same while the composition 
of the available team of agents vary~\cite{Gunn13}. In some other cases, there is historical data present in the form of demonstrations.
This section discusses how to leverage the interactions of previous teams with the environment or past demonstrations to maximize the sum of task rewards in minimal iterations for a new team of agents. 
This translates our problem into the \textit{learning from demonstrations} paradigm. It is  crucial to note that we assume 
a mixture of demonstrations that yield low to high rewards. 
This means that CMTAB is not restricted to the requirement of high-scoring demonstrations.



To process the $n^{th}$ demonstration, CMTAB computes the \textit{task-trait} matrix $Y^{(n)}$ from $X^{(n)}$ and $Q^{(n)}$ as in Eq. (\ref{eq:tasktrait}). Instead of using the assignment $X^{(n)}$ of agents, we use the associated trait distribution $Y^{(n)}$ to learn the reward functions. As a result, even demonstrations of low quality act as \textit{supervisory signals} and can be used to learn valuable trait-reward maps that are generalizable to a new team of agents. Specifically, we generate prior distributions over the trait-reward maps $\{f_m\}_{m=1}^M$ from trait aggregation - reward pairs ($\{y_m^{(n)},\ r_m^{(n)}\}$). These prior distributions help warm-start the online learning process described in Section \ref{subsec:CMTAB}

\section{Experimental Evaluation}

In this section, we describe how we evaluated CMTAB's efficacy in maximizing unknown task rewards using detailed numerical simulations and a simulated emergency response domain. We also conducted experiments evaluating CMTAB's ability to utilize historical data to bootstrap interaction-based learning, and they are discussed in Appendix B.

\vspace{3pt}
\noindent \textbf{Baselines}:
\textcolor{black}{Multi-armed bandits (MAB) represent a natural strategy to tackle the COCOA problem. 
However, as noted earlier, conventional approaches to MAB do not collectively possess all the characteristics that we argue as necessary to effectively solve COCOA. 
As such, we designed baselines that represent conventional MAB techniques. These baselines provide valuable insights into the effectiveness of each component of the CMTAB algorithm.} For all baselines, once the arms were selected at each iteration, we used Equations (\ref{eq:opt})-(\ref{eq:const2}) to identify the coalitions to deploy. 

\begin{enumerate}
\item \textit{Fixed discretization (FD):} This baseline questions the need for adaptive discretization, by uniformly discretizing that continuous trait space $\mathcal{Y}$ at fixed locations. 
In CMTAB, the discretization intervals adaptively change with iterations, whereas, here they are held constant throughout the learning process. 




\item \textit{Adaptive discretization for individual tasks (IA):}
This baseline adaptively determines how to discretize the continuous trait space $\mathcal{Y}$, but does it \textit{separately} for each task.
This baseline challenges CMTAB's \textit{concurrent} selection of arms for all tasks.

\item \textit{Uniform Sampling (US):}
Here, an aggregated trait vector for each task is selected by uniformly sampling \textit{at random} from the continuous trait space $\mathcal{Y}_D$. 
This baseline represents pure exploration, and challenges the need for CMTAB's balance between exploration and exploitation.

\end{enumerate}

\vspace{3pt}
\noindent \textbf{Metrics}:
We measure the performance of CMTAB and that of the baselines using the following metrics:

\begin{enumerate}   
    \item \textit{Best Uncovered Reward} (BUR): Highest expected \textit{ground-truth} total reward uncovered until the current iteration $i$ ($r_{best}^{(i)} = \max_{j\in{\{1,\cdots,i\}}} r_{total}^{(j)}$).
    BUR helps us evaluate effectiveness of exploration and optimization efficiency by measuring how quickly each method is able to identify high-reward regions. 
    \item \textit{Cumulative Multi-task Regret (CMR):}
     Instantaneous multi-task regret at iteration $i$ is defined as the difference between  
     the ground-truth optimal total reward ($r^*_{total} = \max_{[y_1,\cdots,y_m]}\ r_{total} ([y_1,\cdots,y_m])$) for the given team and the total rewards obtained by the allocation sampled at iteration $i$ ($r_{total}(y_m^{(i)})$). As such, CMR at iteration $i$ can be computed as $\bar{R}^{(i)}_{total} = \sum_{j=1}^{i} r^*_{total} - r_{total}(y_m^{(j)})$. CMR evaluates the effectiveness of exploitation by measuring how the suboptimality of sampling accumulates over iterations.
\end{enumerate}

\subsection{Numerical Simulations}
We first evaluated CMTAB on numerical simulations involving different 
reward functions, teams, traits, and hyperparameters. 
In this section, we present the 
results for one set of reward functions and six random teams.

\begin{figure}[t]
    \centering
    \includegraphics[width=0.9\columnwidth]{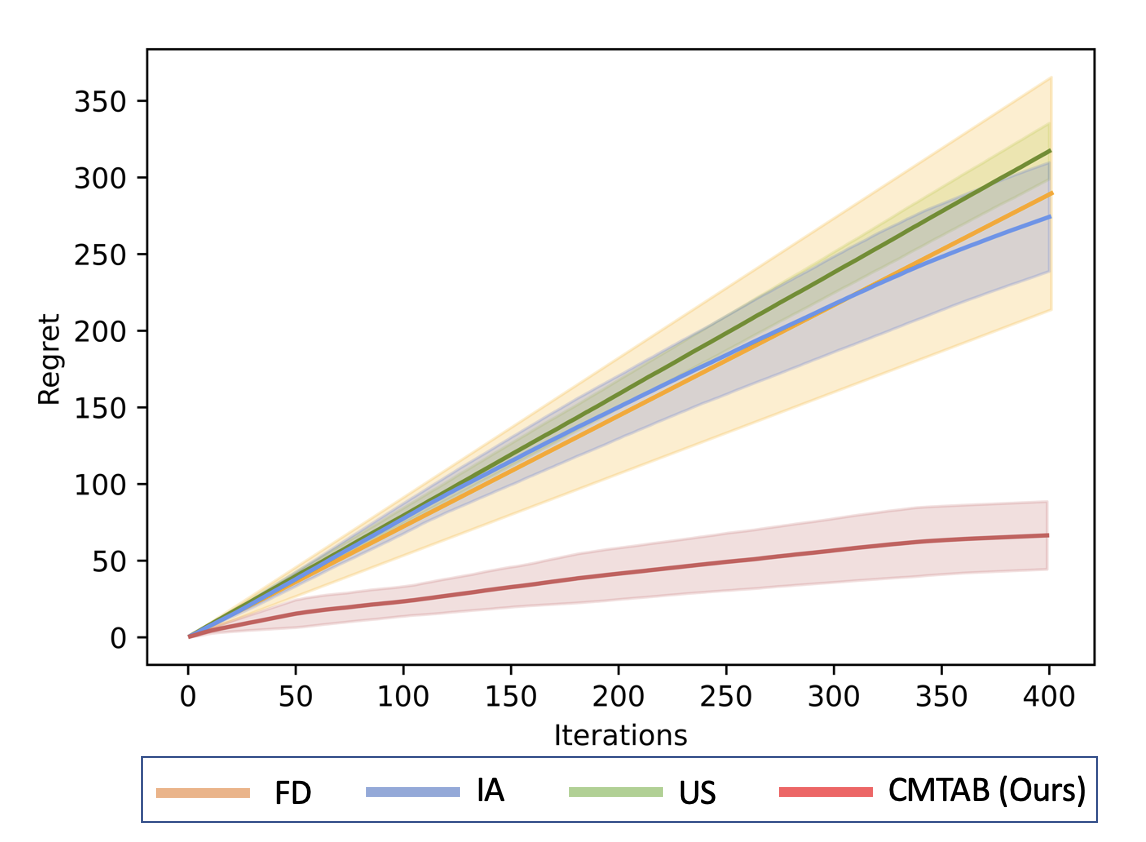}
    \caption{\small{Cumulative multi-task regret averaged over all teams for each baseline when the reward functions are initialized with zero mean}}
    \label{fig:regret_no_prior}
\end{figure}


\vspace{3pt}
\noindent \textbf{Design}

\noindent We generated six instances of the COCOA problems, each with a unique team. Each problem has three tasks ($M$ = 3), four species ($S$ = 4) of robots, and three traits ($U$ = 3). We generated different teams by uniformly randomly sampling the number of robots of each species $N_s$ between 1 and 10. We also randomly generated the trait vector for each species $q_s$ 
(see Appendix A for specific details). To ensure that careful optimization was necessary, we designed these numbers such that the teams had sufficiently different competencies and that they could potentially achieve a wide range of rewards, depending on the quality of allocation.
\textcolor{black}{To investigate CMTAB's ability to handle a wide variety of reward functions, we sampled instances of GPs with randomly generated mean and covariance functions to define the ground-truth reward functions.} Note that the ground-truth reward functions are only used to calculate the BUR and CMR metrics, and are obfuscated from the learning algorithms.
We allowed $N$ = 400 iterations of optimization. Since the environment is assumed to be stochastic, we conducted five rounds of this experiment on each team. The teams are numerically arranged from 1 to 6 in increasing order of their competence, as measured by their respective $r^*_{total}$.


\vspace{3pt}
\noindent \textbf{Results}

\noindent \textit{Exploration efficiency}: In Fig. (\ref{fig:no_prior}), we plot the Best Uncovered Reward (BUR) for all the baselines as a function of number of iterations. We normalized the rewards with respect to the optimal total reward $r^*_{total}$ for the respective team. We see that CMTAB not only outperforms all the baselines but also attains the highest steady-state in the fewest iterations across all six problems. On average, CMTAB achieves 95.5\% of the optimal total reward $r^*_{total}$, compared to 77\% for IA, 68.16\% for US, and 29.1\% for FD. These results suggest that ignoring even one of concurrency (IA), adaptive discretization (FD), or active learning (US) can lead to less efficient optimization.
In particular, we find that fixed discretization (FD) almost completely fails to explore the continuous trait space thus, almost never improving its BUR beyond the first iteration. 
The approximately flat trend in CMTAB's BUR for Team 4 is likely because 
CMTAB discovered a high-reward allocation early, and was unable to find better ones later on.

\vspace{3pt}
\noindent \textit{Cost of exploration}: To quantify the suboptimality incurred by each method in its efforts to optimize the unknown reward functions, we plot the Cumulative Multi-task Regret (CMR) averaged over all the teams in Fig. (\ref{fig:regret_no_prior}). As can be seen, CMTAB accumulates substantially lower regret compared to the baselines. Taken together with the BUR results, this observation suggests that CMTAB maximizes the unknown rewards consistently faster than all baselines while sampling the least suboptimal allocations.

\subsection{Robotarium simulations for Emergency Response Tasks}
\begin{figure}
    \centering
    \includegraphics[width=0.9\columnwidth]{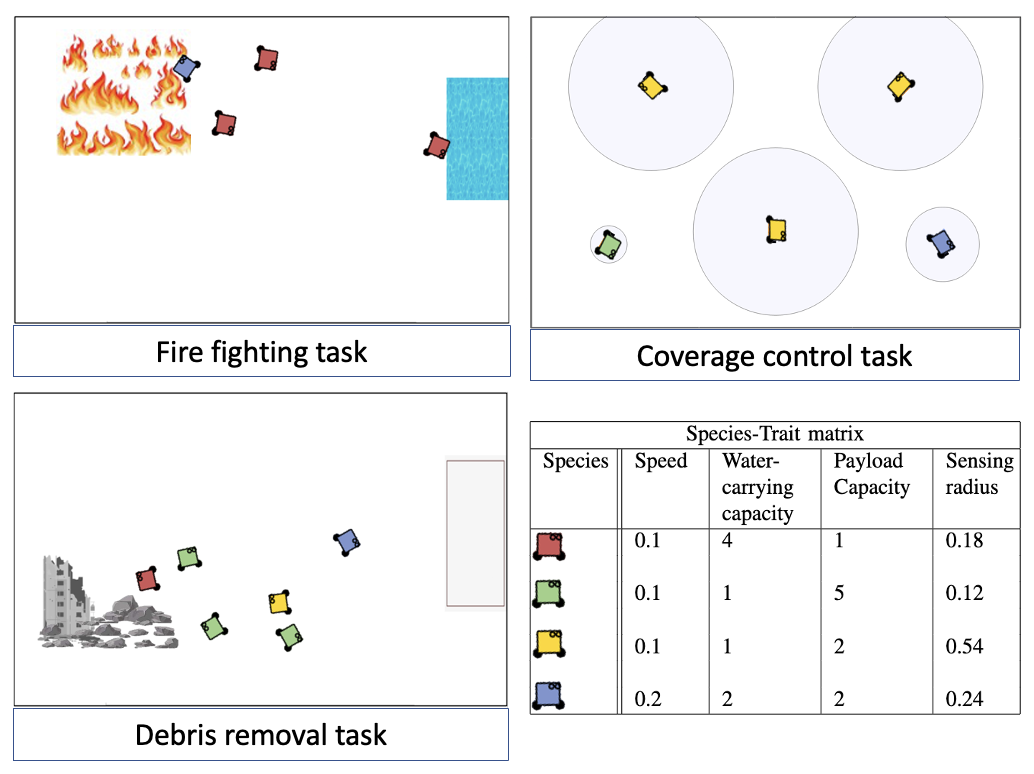}
    \caption{\small{Emergency response tasks in the Robotarium simulator}}
    \label{fig:robotarium_tasks}
\end{figure}
In these experiments, we evaluate the utility of CMTAB in optimizing allocations grounded multi-robot problems. 

\vspace{3pt}
\noindent \textbf{Design}

\noindent We developed an emergency response scenario using the Robotarium simulator \cite{pickem2017robotarium,wilson2020robotarium}.
Our scenario involves three tasks: fire fighting, debris removal, and coverage control for mapping as shown in Fig.~\ref{fig:robotarium_tasks}. All three tasks occur concurrently in the same area, \textcolor{black}{but we have illustrated them separately in the interest of clarity.} We designed a heterogeneous multi-robot team for this scenario with 4 species (S=4), each containing 3-6 robots. 
Each robot species is characterized by 4 traits (U=4): Speed, water-carrying capacity, payload capacity, and sensing radius (see Fig. (\ref{fig:robotarium_tasks}) and the Appendix A for further details).
\textcolor{black}{Once an allocation is identified to sample, we rely on the Robotarium simulator to generate rewards. Specifically, the performance of coalitions in the simulator determines the rewards.}
In this section, we evaluate the performance of CMTAB and the other baselines for one such team of robots (see Appendix C for results involving more teams). 
We executed the Robotarium simulations for $N=400$ iterations for each baseline and repeated this experiment for five rounds. 

\vspace{3pt}
\noindent \textbf{Results}

\begin{figure}%
\centering
\begin{subfigure}{.4\textwidth}
\includegraphics[width=\columnwidth]{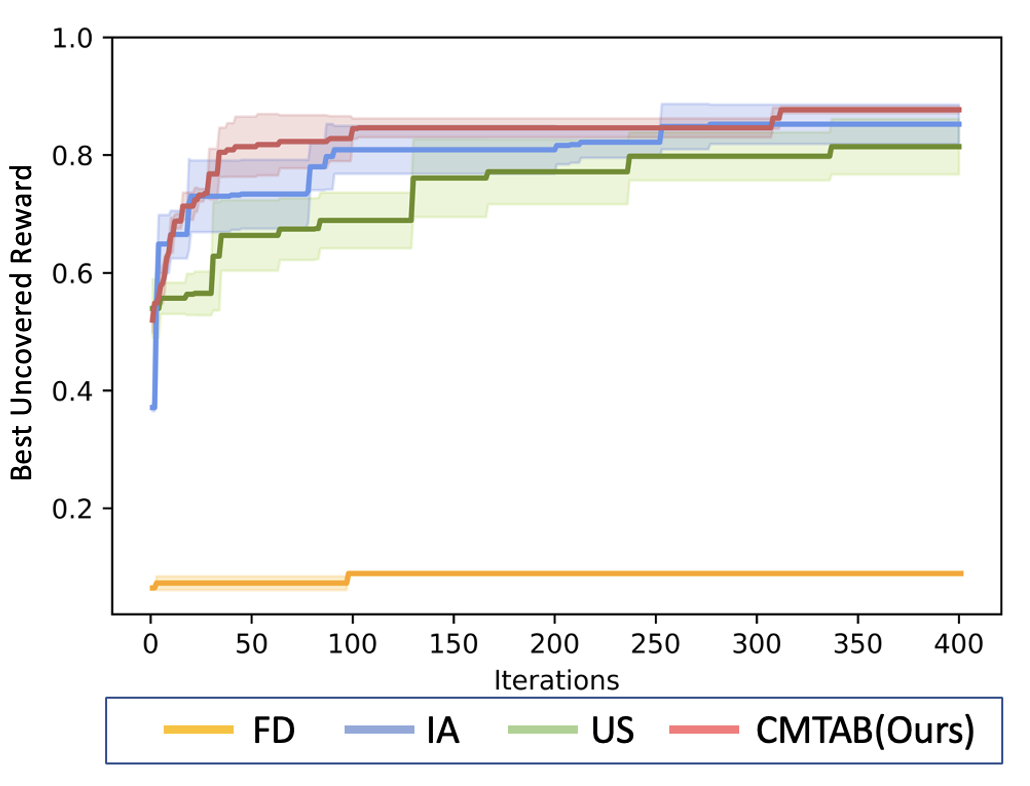}%
\label{fig:robotarium_no_demos_BUR}%
\end{subfigure}%
\\
\begin{subfigure}{.4\textwidth}
\includegraphics[width=\columnwidth]{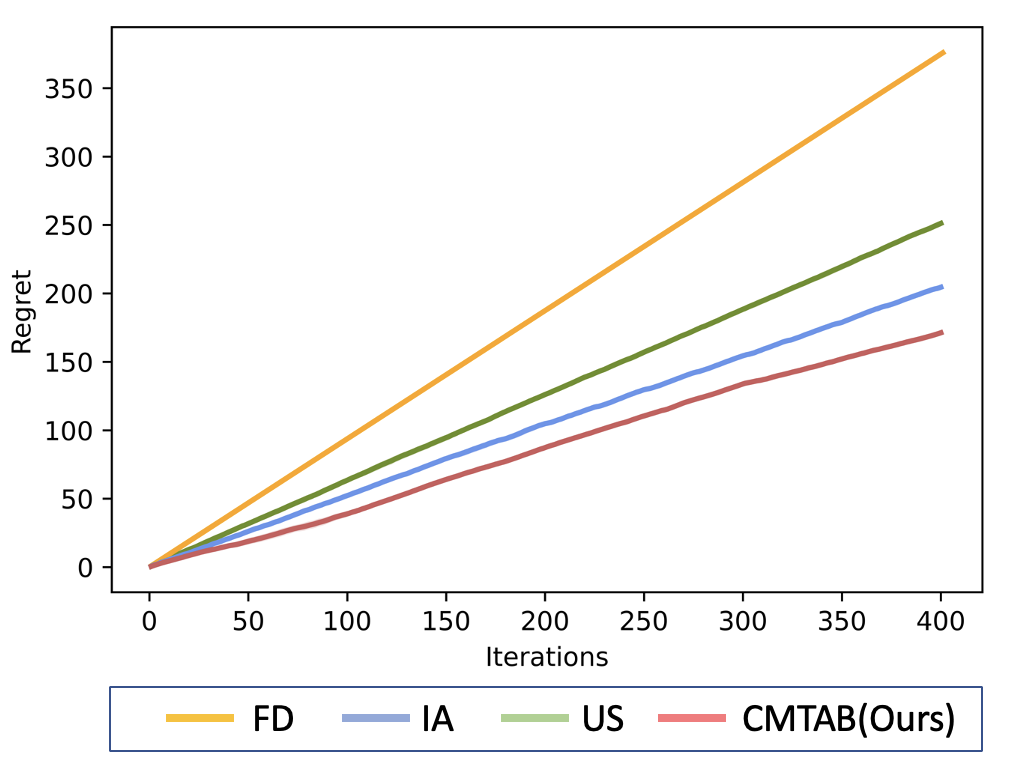}%
\label{fig:robotarium_no_demos_CMR}%
\end{subfigure}%
\caption{Best uncovered reward (top) and Cumulative multi-task regret (bottom) over iterations in the simulated emergency response environment.}
\label{fig:robotarium_results}
\end{figure}

\noindent As seen in the BUR and CMR plots of Fig.~(\ref{fig:robotarium_results}),
CMTAB consistently outperforms the FD and US baselines and marginally better than the IA baseline both in terms of BUR and CMR. On average, the IA and CMTAB baselines reach 95.95\% of their highest BUR in the first 100 iterations. These results highlight the importance of \textit{adaptive discretization}, an approach common in both, the IA and CMTAB baselines. The FD baseline's poor performance is likely explained by the possibility that the fixed discretization points on the trait aggregate space are not close to the points that are achievable by the team. As such, FD could not identify high-reward regions.   

\section{Conclusion}
We formulate a new class of problems, dubbed \textit{COCOA}, that involves online optimization of heterogeneous multi-robot task allocation when the utility or reward functions are not known. We also contribute a novel bandit-based algorithm, named \textit{CMTAB}, to solve the COCOA problem by addressing two of its central challenges (task concurrency and resource constraints) by carefully trading off exploration and exploitation.
Our detailed experiments involving both numerical simulations and a simulated emergency response domain reveal that the three main design choices of CMTAB (trait-reward maps, adaptive discretization of continuous trait space, and concurrent sampling) are essential to efficiently maximize the unknown reward functions. Specifically, we show that CMTAB can identify and exploit high-performing allocations in as few iterations as possible, reducing the cost of exploration as measured by accumulated regret.
We also demonstrated that the CMTAB can leverage historical data as demonstrations to bootstrap online learning, even when the demonstrations are suboptimal. 
While empirical results show that CMTAB had the lowest multi-task regret compared to baselines, future work can derive theoretical bounds on CMTAB's regret.
\textcolor{black}{Scalability with respect to the number of traits and tasks is another avenue for further investigation.}

%



\bibliographystyle{plainnat}
\bibliography{references}

\end{document}